\documentclass[preprint,review,12pt]{elsarticle}

\usepackage{amssymb}
\usepackage{graphicx}
\usepackage{caption}
\usepackage{subcaption}
\usepackage{amsmath}
\journal{Neural Networks}

\begin{document}

\begin{frontmatter}

%% Title, authors and addresses

%% use the tnoteref command within \title for footnotes;
%% use the tnotetext command for theassociated footnote;
%% use the fnref command within \author or \address for footnotes;
%% use the fntext command for theassociated footnote;
%% use the corref command within \author for corresponding author footnotes;
%% use the cortext command for theassociated footnote;
%% use the ead command for the email address,
%% and the form \ead[url] for the home page:
%% \title{Title\tnoteref{label1}}
%% \tnotetext[label1]{}
%% \author{Name\corref{cor1}\fnref{label2}}
%% \ead{email address}
%% \ead[url]{home page}
%% \fntext[label2]{}
%% \cortext[cor1]{}
%% \address{Address\fnref{label3}}
%% \fntext[label3]{}

\title{Control of the Correlation of Spontaneous Neuron Activity in Biological and Noise--activated CMOS Artificial Neural Microcircuits}

%% use optional labels to link authors explicitly to addresses:
%% \author[label1,label2]{}
%% \address[label1]{}
%% \address[label2]{}

\author[Addr1,Addr2]{Ramin M. Hasani}
\author[Addr1]{Giorgio Ferrari}
\author[Addr3]{Hideaki Yamamoto}
\author[Addr4]{Sho Kono}
\author[Addr4]{Koji Ishihara}
\author[Addr4]{Soya Fujimori}
\author[Addr5]{Takashi Tanii}
\author[Addr6]{Enrico Prati}

\address[Addr1]{Dipartimento di Elettronica, Informazione e Bioingegneria, Politecnico di Milano, Via Colombo 81, I--20133 Milano, Italy}
\address[Addr2]{Department of Computer Engineering, Vienna University of Technology, Treitlstrasse 3-3, 1040, Vienna, Austria}
\address[Addr3]{Frontier Research Institute for Interdisciplinary Sciences, Tohoku University, 6-3 Aramakiaza-Aoba, Aoba, Sendai, Miyagi 980-8578, Japan}
\address[Addr4]{Department of Electronic and Physical Systems, Waseda University, 3-4-1 Ohkubo, Shinjuku, Tokyo 169-8555, Japan}
\address[Addr5]{School of Science and Engineering, Waseda University, 3-4-1 Ohkubo, Shinjuku, Tokyo 169-8555, Japan}
\address[Addr6]{Istituto di Fotonica e Nanotecnologie, Consiglio Nazionale delle Ricerche, Piazza Leonardo da Vinci 32, I--20133 Milano, Italy}

\begin{abstract}
There are several indications that brain is organized not on a basis of individual unreliable neurons, but on a micro-circuital scale providing Lego blocks employed to create complex architectures. At such an intermediate scale, the firing activity in the microcircuits is governed by collective effects emerging by the background noise soliciting spontaneous firing, the degree of mutual connections between the neurons, and the topology of the connections. We compare spontaneous firing activity of small populations of neurons adhering to an engineered scaffold with simulations of biologically plausible CMOS artificial neuron populations whose spontaneous activity is ignited by tailored background noise. We provide a full set of flexible and low-power consuming silicon blocks including neurons, excitatory and inhibitory synapses, and both white and pink noise generators for spontaneous firing activation. We achieve a comparable degree of correlation of the firing activity of the biological neurons by controlling the kind and the number of connection among the silicon neurons. The correlation between groups of neurons, organized as a ring of four distinct populations connected by the equivalent of interneurons, is triggered more effectively by adding multiple synapses to the connections than increasing the number of independent point-to-point connections. The comparison between the biological and the artificial systems suggests that a considerable number of synapses is active also in biological populations adhering to engineered scaffolds.
\end{abstract}

\begin{keyword}
%% keywords here, in the form: keyword \sep keyword
Cortical microcircuits \sep
Tonic spiking \sep
Silicon brains \sep
Neuromorphic engineering \sep
Silicon neurons \sep
Learning in silicon \sep
Noise injection \sep
Covariance analysis \sep

%% PACS codes here, in the form: \PACS code \sep code

%% MSC codes here, in the form: \MSC code \sep code
%% or \MSC[2008] code \sep code (2000 is the default)

\end{keyword}

\end{frontmatter}

%% \linenumbers

%% main text
\section{Introduction}
\label{S1}
%<Introduction>
%	<Objective of the work and why it is important (starting sentence)>
The correlation of biological and noise-solicited CMOS artificial spiking neurons between small distinct populations of neurons is modulated and compared by controlling the number of connections and of synapses.
%	<Objectives of the work>
The exploration of neuronal microcircuits \cite{buldyrev2000description,jones2000microcolumns,haeusler2007statistical,haeusler2009motif,rinkus2010cortical,markram2011innate,potjans2014cell,opris2014prefrontal,wang2016brain}
as Lego building blocks at the basis of complex architectures is not only supported by experimental evidence \cite{sadovsky2014mouse,ocker2015self} but also substantiated by mathematical models \cite{xie2016brain}.

The study of biological neurons at few neurons scale represents an unavoidable intermediate step towards the comprehension and the control of building blocks of efficiently engineered artificial neural networks. Minimal circuits of two neurons \cite{bengtsson2013cross}, including key ingredients like activation with noise \cite{zeldenrust2013modulation} have been recently explored.

The major role of noise as neural computation resource has been recognized
long ago in both biological  \cite{burton1992event, lukashin1996modeling,nozaki1999effects,natschlager2005dynamics,brascamp2006time,ecker2014there, orlandi2013noise,vukovic2015robust,audhkhasi2016noise} and artificial \cite{hinton1983optimal,ackley1985learning, lizeth2012impact, prati2016noise, prati2016atomic} neural networks. Experiments on a random network of spiking neurons show that a significant amount of knowledge is stored stochastically inside the nervous system \cite{habenschuss2013stochastic} and applications of noise in the learning process inside a spiking neural network (SNN) have been discussed \cite{maass2014noise}.

\begin{figure}[!ht]
\centering
\includegraphics[width=1\textwidth]{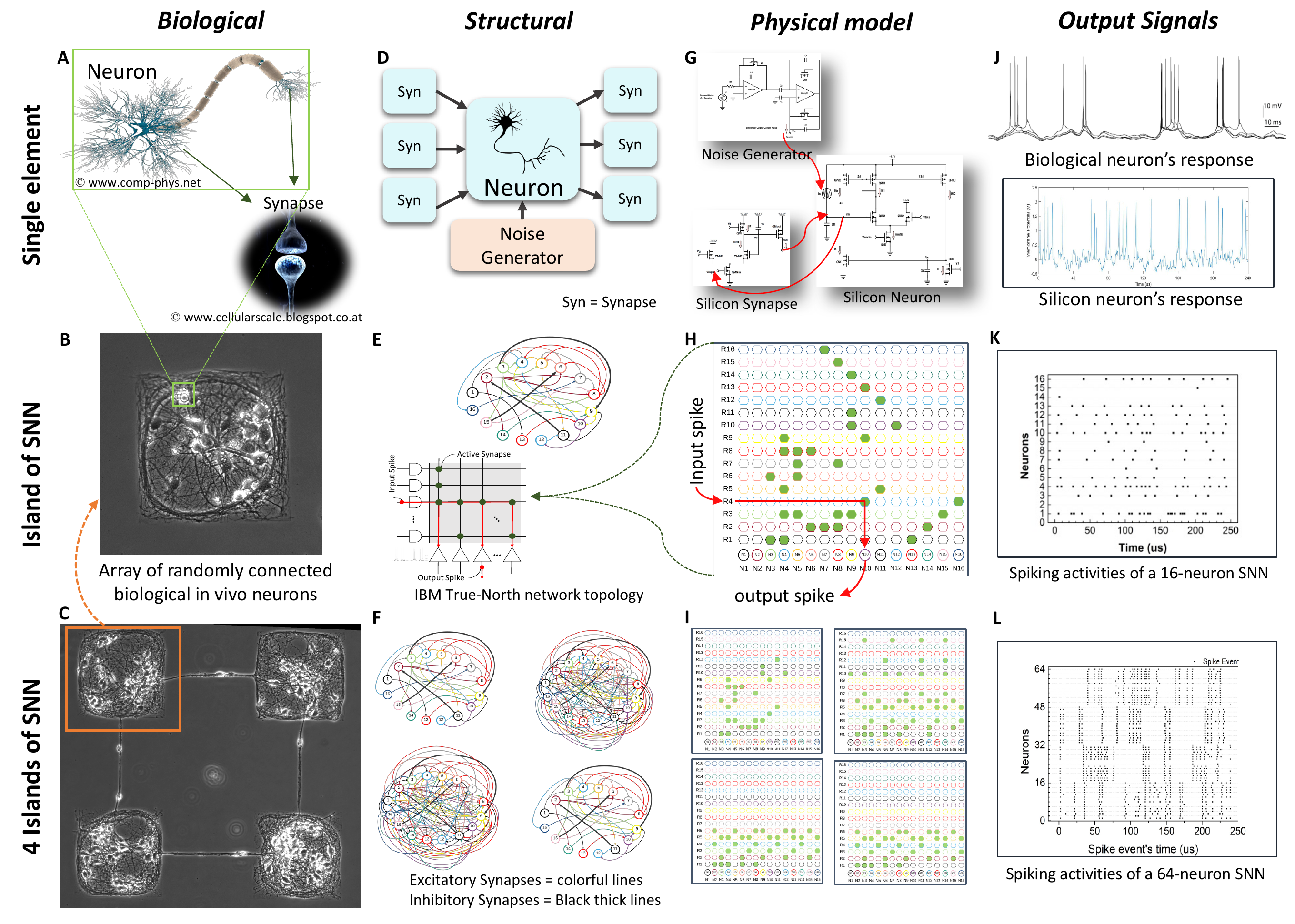}
\captionsetup{width=1.5\linewidth}
\caption{General framework for modeling and exploiting noise induced spiking activity from biological to silicon neurons. Anatomy of the elements of a neural network including neurons and synapses, together with their neural network structure, is shown in biological, structural and physical model representation. A) Anatomy of a single neuron with a zoom in on synapses region B) A network of in-vitro neurons in a random topology connectome. C) Four islands of biological neurons linked to each other by a single interneuron where we observe the propagation of the correlation of spiking activity of single neurons by increasing and decreasing the number of interneuron connections. D) Structural representation of a stochastic model of spiking neuron and chemical synapses. E) Network of 16 neurons randomly connected to each other. We designed our network inspired by the structural topology used by IBM TrueNorth \cite{merolla2014million}. F) Structure of four island of randomly connected CMOS spiking neurons isolated form each other. We simulate a similar experiment performed on real neurons by CMOS SNNs. G) Stochastic model of spiking neuron designed in a 0.35 $\mu m$ CMOS technology together with the designed CMOS synapses. H) Network of 16 CMOS neurons and 256 synapses designed in a 0.35 $\mu m$ CMOS technology. The network is designed such that any arbitrarily selected topology can be implemented on it. I) 4 island of CMOS neurons and synapses. The network consist of 64 neurons and 1024 synapses. J) output response of a biological and a CMOS neuron to Gaussian white-noise signal. K) Raster plot of the spiking activity of single neurons of the network of (H), where a Gaussian white-noise source excites one single neuron of the network. L) Raster plot of the spiking activity of 64 neurons of the network of (I), where neurons of each island are excited by an independent source of white-noise.}
\label{f1}
\end{figure}

%	<Justification of the objectives>
%	<Background>
Networks of stochastic spiking neurons have been used for solving constraint satisfaction problems where noise is believed to play a key role in the problem solving ability of the human brain \cite{maass2015spike, jonke2016solving, kappel2015network, mostafa2015event}. %		How?
For instance, the individual activity of neurons are correlated in the isolated regions of auditory cortex to which the neuron belongs \cite{nir2007coupling}. The correlation among distinct groups of neurons (\textit{islands }for brevity from now on) are directly proportional to the number of interneuron connections and interneuronal correlation. Furthermore, synchronization among neuronal populations in motor cortical regions where they communicate through interneurons, may play an important role in cognitive motor processes \cite{riehle1997spike}.
Propagation of spiking activity amongst different neuronal modules has been reviewed in \cite{kumar2010spiking} Refinements of a natural communication stimulus is permitted by the correlated neural activity in fish \cite{metzen2016neural}.  A mechanism by which a neural circuits effectively shape their signal and noise in concert has been demonstrated, enabling the minimization of the corruption of signal by noise and by enhancing the speed of information transmission \cite{zylberberg2016direction}.

%		What have WE done previously?
Previously, we separately explored the island size-dependence of synchronized activity in living neuronal networks \cite{yamamoto2016size} and the Gaussian white-noise to enhance the transmission of spiking activity along a linear chain of artificial neurons based on discrete components, and its inter-spike interval (ISI) by varying the amplitude of the noise \cite{prati2016noise}.

%	<Guidance to the reader>
%		What should the reader watch for in the paper?
%		What are the interesting high points?
%		What strategy did we use?
CMOS artificial neurons \cite{cassidy2013design,sharifipoor2012analog,yasukawa2016real}
 represent the ideal platform to emulate biologically plausible neurons operating in speed mode and implementing universal functionalities speculated for Lego block microcircuits and build engineered artificial neural networks.

Here we exploit the effects of electronic noise in a network of simulated CMOS artificial spiking neurons in order to reverse engineer the features of a SNN of real neurons attached to a silicon patch.
%	<Summary/Conclusion>
%		What should the reader expect as conclusion?
%		Details of the sections of the experimental results
%<\Introduction>
Figure \ref{f1} graphically represents an overview of our research framework. With the goal of observing the change of correlation rate of spiking activity of neurons among different islands of neurons by adding interconnection neurons, we performed experiments with real in vitro neurons, as the first part of this research. Spiking activity of four islands of randomly connected spiking neurons with sufficiently interconnected neurons has been investigated. We observed that by increasing the number of connections between islands the correlation of spiking activities dramatically rises up.
In the second part, we aim to the exploitation of a nature inspired physical effect such an electric noise (by adopting the same network topology of the first part) on silicon neurons. It is therefore divided as follows: we initially describe the network elements (a CMOS neuron operating in speed mode, both excitatory and inhibitory silicon synapse and two compact noise generators for both white and pink noise), briefly. Such components are combined to build up a network of spiking neuron ready to investigate the effects of noise on it. The noise response of the silicon neuron to different noise spectra is discussed. 
The silicon neuron microcircuits are utilized to investigate the correlation of noise assisted spiking activity of the four island topology. As key results, we demonstrate how correlation of spiking activity among isolated sub-networks ignited by distinct background noise is progressively increased by adding interconnection elements between neurons of different sub-networks, and more efficiently by adopting multiple synapses ensuring firing to more input neurons, by achieving the artificial microcircuit counterpart of the biological network results. 

\section{Spontaneous firing activity of group of biological neurons}
Biological neurons, which are obtained from animal brains \cite{kaech2006culturing} or stem cell differentiation \cite{shi2012directed}, can be cultivated in vitro under a defined physicochemical condition. Neurons are adherent cells, and hence an appropriate scaffold is essential for their growth. A number of microfabrication methods, including semiconductor lithography or soft lithography, has been utilized to engineer the scaffold to control the growth area at single-cell or multi-cellular scales \cite{wheeler2010designing, aebersold2016brains, yamamoto2016size, yamamoto2016unidirectional, kono2016live}.

After approximately a week in culture, neuronal networks begin to form synaptic connections and spontaneously generate bursting activity that is highly synchronized within the network \cite{yamamoto2016size, orlandi2013noise}. We previously showed using micropatterned rat cortical neurons that neuronal correlation in the spontaneous bursting activity decreases with the size of the network \cite{yamamoto2016size}. In this section, we use a similar culture to show that correlation between a group of living neurons can be regulated by altering the number of neurite guidance pathways that bridge neuron islands.

The materials and methods used in fabricating micropatterned substrates, culturing primary neurons, and recording spontaneous neural activity have been described previously \cite{yamamoto2016size, kono2016live}. Briefly, $poly-D-lysine$ and $2-\lbrack$\textit{methoxy (polyethleneoxy)propyl}$\rbrack$\textit{trimethoxysilane} were used as cell-permissive and non-permissive layers, respectively, and were patterned on a glass coverslip using electron-beam lithography. The micropatterns used in the current experiment consisted of a set of four square islands of $200\times200$ $\mu m^2$ and $5\mu m$-wide lines that interconnect the island. Three geometries with different number of interconnecting lines were compared, i.e., no-bond and triple-bond structures which have zero and three lines between a pair of islands.

Primary neurons were obtained from embryonic rat cortices, plated on the micropatterned coverslips, and cultured for 10 days. Then, the cells were loaded with the fluorescence calcium indicator Fluo-4 AM (Molecular Probes), and their spontaneous activity was measured by calcium imaging. Images were obtained every $200 ms$ for 6 or 9 min using an inverted microscope (Nikon Eclipse TE300) equipped with a $20\times$ objective lens (numerical aperture, 0.75) and a cooled-CCD camera (Hamamatsu Orca-ER). The image sequences were later analyzed offline using the ImageJ (NIH) and custom-written Perl programs \cite{yamamoto2016size}.

Representative phase-contrast micrographs of living neuronal networks grown on the micropatterned substrate are shown in Figure \ref{f1}B and \ref{f1}C . Apparently, neurons adhered and grew neurites only on the permissive domains. Very little non-specific adhesion or neurite growth was observed on the non-permissive area. The number of cells in each network was counted from the images and were evaluated to be 148, 143, and 154 cells for the no-bond, single-bond, and triple-bond structures, respectively.
\begin{figure}[!ht]
\centering
\includegraphics[width=1\textwidth]{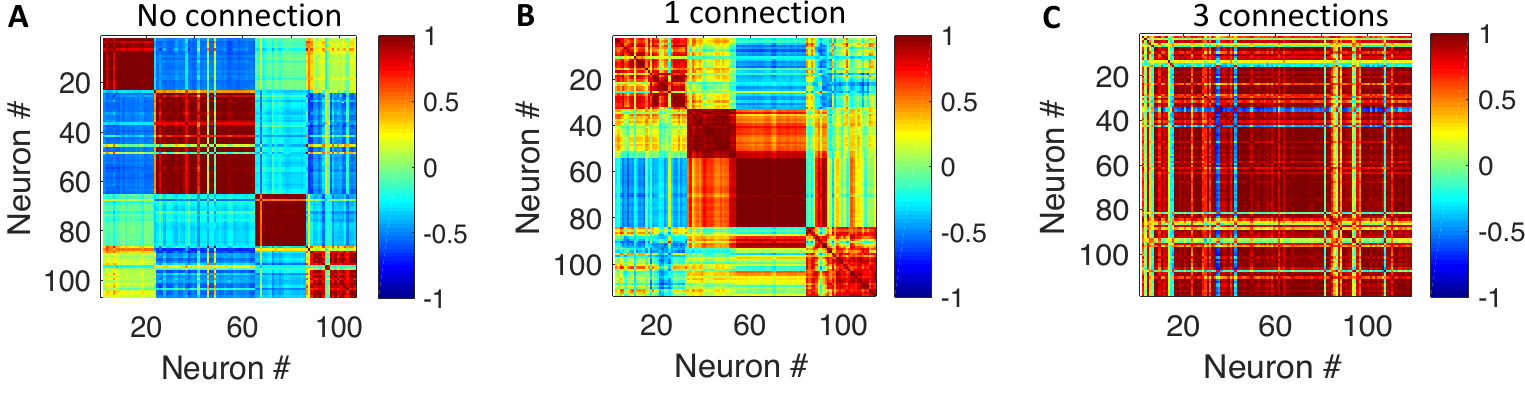}
\captionsetup{width=1.5\linewidth}
\caption{The Pearson correlation coefficient matrix of 4 islands of biological neurons represented in Figure \ref{f1}C. The color-bar shows the amount of correlation coefficient from uncorrelated regions (blue) to highly correlated areas (dark red). In the case A the islands are not connected and the correlation is significant only for neurons within the same island. In case B and C, the correlation progressively increases by adding 1 and 3 neuron connections respectively among the islands in a ring topology.}
\label{f02}
\end{figure}

Next, spontaneous neural activity was measured by fluorescence calcium imaging, and the effect of changing the number of interconnections between neuronal islands was analyzed by evaluating the Pearson correlation coefficient. Correlation coefficient of neuron pairs A and B, is calculated using the Pearson correlation coefficient equation described in Appendix section. The correlation coefficient matrix is calculated for three different situations and it is reported in Figure \ref{f02}. In the no-bond structure, the correlation of the synchronized bursting activity was high only for neurons belonging to the same island and were otherwise nearly zero. By connecting the islands of biological neurons by an individual connection, the correlation among different islands is switched on. Finally, in the triple-bond structure, the correlation is high across nearly all neuron pairs. %The dynamics of the single-bond structure was in an intermediate regime.

The results indicate that the neural correlation can be controlled using micropatterned substrates by changing the number of interconnections. This is in agreement with a previous work which showed that the likelihood of an activity being transferred between two neuronal populations increased with the number of interconnecting microtunnels in a microfluidic device \cite{pan2015vitro}. Detailed statistical analysis of the structure-dependent modulation in the activity pattern is in progress and will be reported elsewhere soon.

The spontaneous activity in biological neuronal networks is triggered by spontaneous release of either synaptic vesicles or ion channel stochasticity which temporally induce fluctuation of the  neuronal membrane potential. It is interesting from an engineering point of view that the up to 80\% of the metabolic energy consumed in the brain seems to be used in maintaining the spontaneous activity \cite{raichle2006brain}. Their possible role in neural computation has been suggested to include learning \cite{burton1992event}, event directionality \cite{lukashin1996modeling}, stochastic resonance \cite{nozaki1999effects},
management of binocular rivalry \cite{brascamp2006time}, and coherence \cite{orlandi2013noise}, just to mention some examples, but full understanding of the functional role of such an energy-consuming background activity awaits further research.

\section{Exploiting injected noise in networks of silicon neurons}
\begin{figure}[h!]
\centering
\begin{subfigure}[t]{.5\textwidth}
\centering
\includegraphics[width=2.5in]{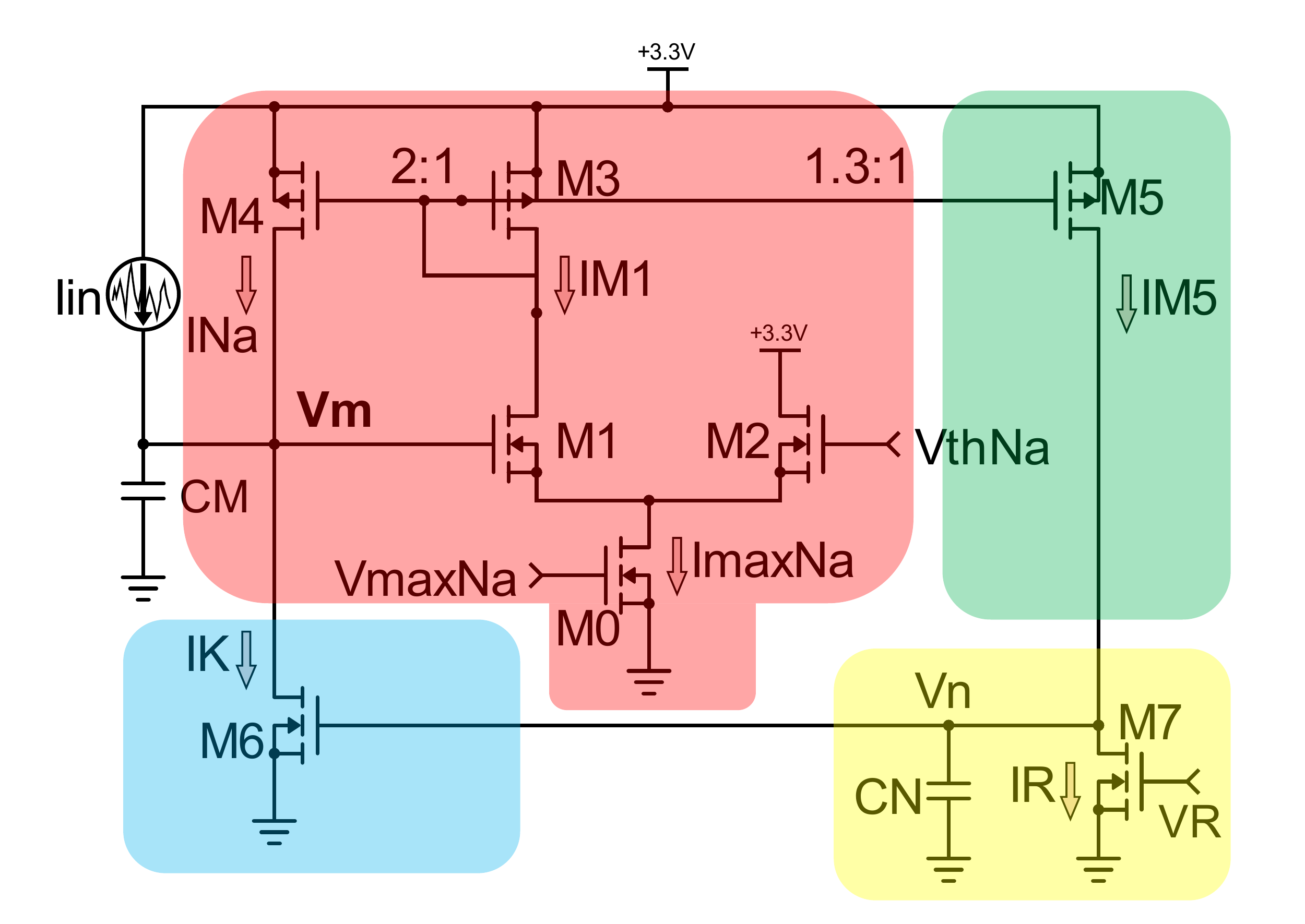}
\caption{Sodium-potassium neuron circuit \cite{sarpeshkar1992refractory}}
\label{elements1}
\end{subfigure}%
\begin{subfigure}[t]{.6\textwidth}
\centering
\includegraphics[width=2in]{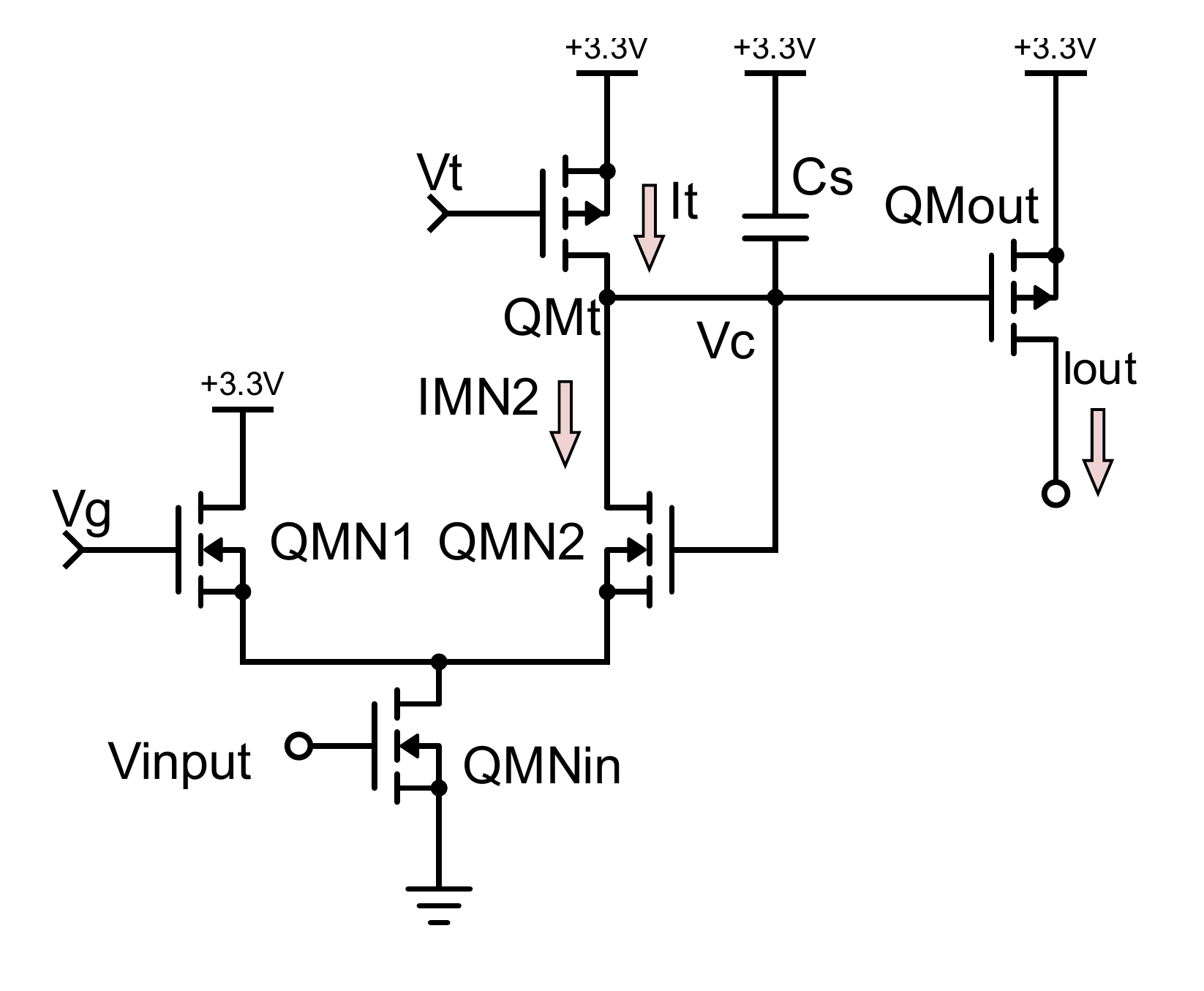}
\caption{CMOS excitatory synapse \cite{chicca2014neuromorphic}}
\label{elements2}
\end{subfigure}
\begin{subfigure}[t]{.4\textwidth}
\centering
\includegraphics[width=2.1in]{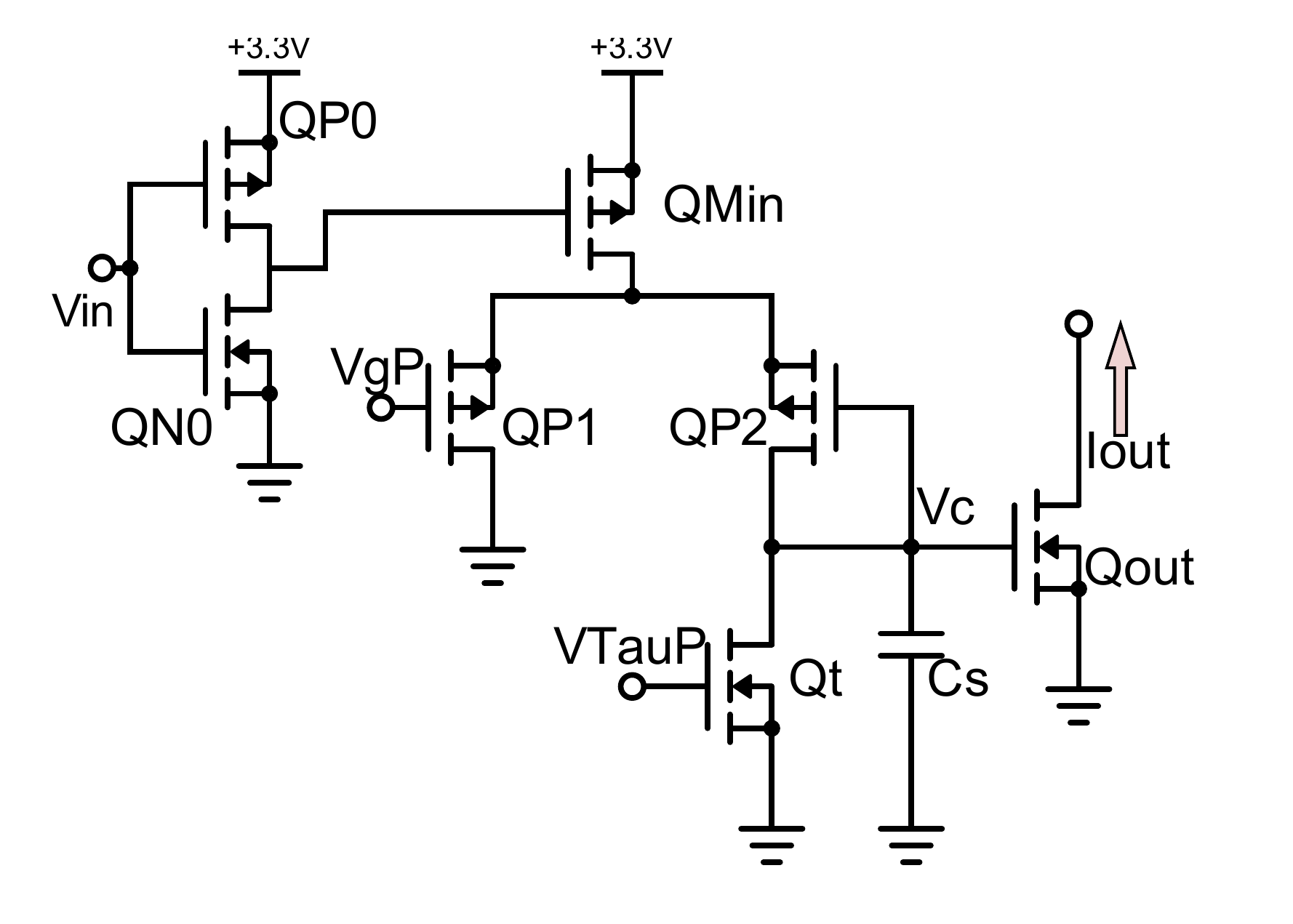}
\caption{CMOS inhibitory synapse }
\label{elements3}
\end{subfigure}
\begin{subfigure}[t]{.45\textwidth}
\centering
\includegraphics[width=2.5in]{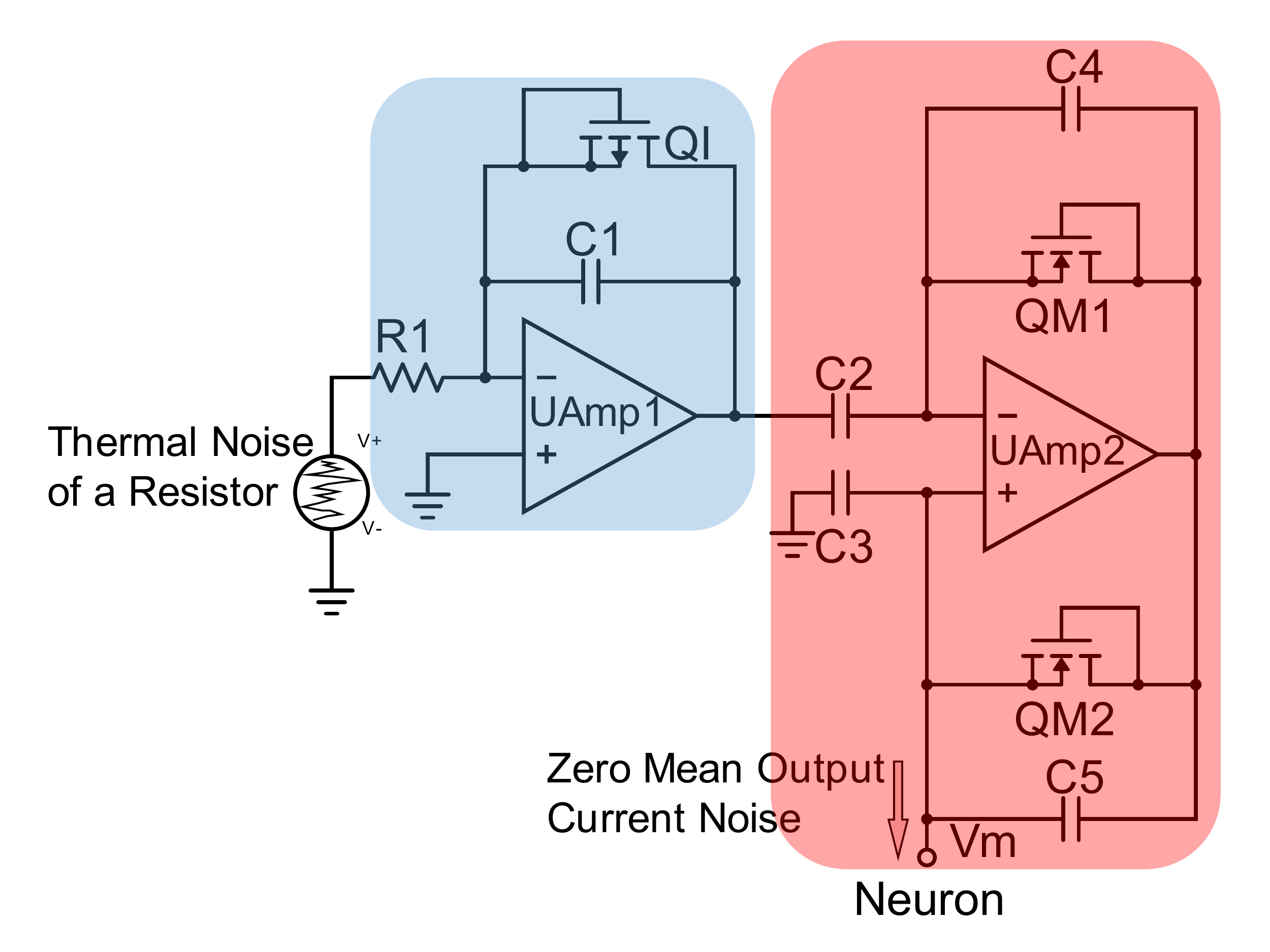}
\caption{Voltage-current converter}
\label{elements4}
\end{subfigure}
\captionsetup{width=1.5\linewidth}
\caption{Schematic of the designed network elements. a) CMOS implementation of a fast-mode sodium-potassium neuron. The neuron circuit comprises a voltage-gated sodium conductance channel (red), a voltage-gated potassium channel (light blue), a coupling element between the sodium and potassium channels (green) and a refractory period tuner (yellow).  The voltage, $V_m$, on capacitor $C_m$ is the membrane potential. Biasing voltages $V_{Na}^{th}$, $V_{Na}^{max}$ and $V_R$ set the threshold for firing, pulse width and refractory period of the action potential, respectively. The value of capacitors together with input current $I_{in}$ determine the frequency of firing of the neuron. b) CMOS implementation of an excitatory Synapse; using a current-mode log-domain CMOS differential-pair integrator (DPI) \cite{chicca2014neuromorphic}. c) Complementary DPI synapse circuit designed and utilized as inhibitory synapse. d)A CMOS voltage to current (V-I) converter with a very low output dc current designed for converting the thermal noise of a resistor into a zero-mean white-noise current. The V-I converter circuit consists of a modified holland current source structure (red) \cite{instruments2008comprehensive} together with an integrator(blue). The current noise is applied as input to each neuron in order to induce stochastic behavior of spiking neurons. Note that thermal noise of a silicon resistor has been used for generating the input noise.}
\label{elements}
\end{figure}

In this section we simulate the implementation of network of CMOS spiking neurons and the effects of injecting noise. Noise is naturally present in CMOS devices, from white and pink ($1/f$) noise \cite{nemirovsky20011}, to telegraph noise \cite{prati2007microwave}, to even $1/f^{1/2}$ noise \cite{prati2016band}. Therefore, the most straightforward strategy consists of amplifying such natural noise and to exploit it in the circuit by injection in the selected nodes.
Selected topology of the network is inspired by the experiments performed on biological neurons. Hudgkin-Huxley (HH) spiking neurons \cite{hodgkin1952quantitative} are designed and simulated in a 0.35 $\mu m$ CMOS technology \cite{sarpeshkar1992refractory}. Figure \ref{elements1} depicts the structure of the silicon neuron designed in sub-threshold regime where the sodium and potassium conductance channels are modeled. Neurons are designed to be able to spike three orders of magnitude faster than real neurons. Speed mode allows fast data processing more importantly it takes advantage of scaling of the size of the circuit. The silicon neurons are mutually connected by means of CMOS excitatory and inhibitory synapses (see Figure \ref{elements2} and \ref{elements3} respectively). With the aim of applying noise to SNNs, a voltage to current converter (Figure \ref{elements4}) has been developed in order to deliver a zero mean white and pink current noise to the neurons. In the following, such basic elements of the network are discussed.

$-$ \textit{\textbf{Fast-mode (HH) Silicon Neuron}}. Circuit of Figure \ref{elements2} realizes a sodium-potassium conductance based model of spiking neurons biased in sub-threshold regime. Capacitor $C_m$ stands for the membrane capacitance. When an input current stimulates the capacitance, it starts to be charged and consequently membrane potential, $V_m$ increases. Therefore transistor M1 turns on and sinks current $I_{M1}$. The current then is mirrored to the transistor M4 and creates the sodium activation current, $I_{Na}$ which builds up the up-swing of the action potential. Simultaneously, $I_{M1}$ is also copied to the transistor M5 and as a result of the so current $I_{M5}$, the capacitor $C_{N}$ get charged and activates the transistor M6, ergo it sinks the current $I_{K}$. Therefore, $C_{m}$ is discharged and $V_{m}$  drops to its resting potential. Threshold of activation of the sodium conductance is set by $V_{Na}^{th}$. It controls the threshold of firing of the neuron. Pulse width of the spike is controlled by $I_{Na}^{max}$ through $V_{Na}^{max}$. Refractory period of the neuron is set by $I_{R}$ set by the voltage $V_{R}$ activating the transistor M7. The neuron is designed such that it fires three orders of magnitude faster than the biological neurons. In such way, respecting the working principles of a real neuron, one can acquire more spikes in shorter simulation times and enables fast processing. The size of a single CMOS neuron is therefore reduced to $35\times35 \mu m^2$, in contrast to previous implementations such as $90\times90 \mu m^2$ in \cite{serrano2006neuromorphic}(more examples in\cite{indiveri2011neuromorphic}).

$-$ \textit{\textbf{Excitatory Silicon Synapse}}. Biological synapses are modeled by an exponentially decaying time course, having different time constants for different types of synapses \cite{destexhe1998kinetic}. Therefore, a first-order differential equation of the type:
\begin{equation}
\label{eq1}
	\tau \dot I = -I + I_{in},
\end{equation}
where $I$ stands for the output current, and $I_{in}$ represents the input stimulus to the synapse, can be implemented in order to model synaptic transmission \cite{destexhe1998kinetic}. We use a current-mode low-pass filter in order to model the synapses. We design the synapse to be compatible with the silicon neuron. Therefore the circuit has much greater bandwidth in comparison to that of biological synapses. A differential-pair Integrator (DPI) circuit \cite{chicca2014neuromorphic} shown in Figure \ref{elements2} is designed as an excitatory synapse. Voltage-spikes from the presynaptic neurons arrive at the input transistor $QMN_{in}$ convert to a current, $I_{in}$. By applying trans-linear principle in which, in the log domain circuits, the sum of transistor voltages can be replaced by the multiplication of their currents \cite{gilbert1996translinear}, and by writing the current-voltage relationship of the capacitor $C_s$, the dynamics of the circuit can be presented as:
\begin{equation}
\label{eq2}
	\tau \frac{d}{dt}I_{out} = -I_{out} + I_{in}  \frac{(\frac{I_{out}}{I_{\tau}})}{1 + (\frac{I_{out}}{I_{\tau}})},
\end{equation}
where the time constant is $\tau = \frac{C_s U_T}{k I_{\tau}}$ in which $C_s$ represents the synapse capacitor, $k$ is the sub-threshold slope factor and $U_T$ stands for the thermal voltage. Note that by adjusting the value of the synapse capacitor, $C_s$, and $I_{\tau}$ one can control the time constant of the synapse.

$-$ \textit{\textbf{Inhibitory Silicon Synapse}}. The inhibitory synapse consists of complementary version of the excitatory synapse shown in Figure \ref{elements2}. Figure \ref{elements3} represents the inhibitory synapse circuit. Working principle of such circuit is similar to that of excitatory synapse. All n-MOS transistors are replaced with p-MOSs and vice versa. At the input of the inhibitory synapse, a CMOS inverter circuit consisting of the transistors indicated as QP0 and QN0, is used to invert the upcoming voltage-spikes from the presynaptic neuron. Eventually, an inhibiting current pulse $I_{out}$ is generated at the output of the circuit once the synapse get stimulated from a presynaptic neuron.

$-$ \textit{\textbf{CMOS Voltage to Current Converter}}. CMOS V-I converter shown in Figure \ref{elements4}, is designed in order to inject noise signal of a silicon resistor to the silicon neuron. 
We use a resistor with larger value in the range of k$\Omega$ and amplify the current in order to deliver the required amplitude of the noise. An integrator (negative-feedback inverting structured integrator using opamp UAmp1) together with a capacitor, C2, is utilized for current amplification purpose. If we connect the capacitor, C2, directly to the neuron, the structure will provide a current amplification with the ratio (C2/C1) and it is AC coupled. The main disadvantage is that the output current depends on the input voltage of the neuron therefore the system is not a true current source. Moreover, we need a zero-mean thermal noise to be applied to the neuron. As a result, we add the second stage by utilizing a modified Howland current source structure \cite{instruments2008comprehensive} shown in the red box in Figure \ref{elements4}. At the output of a current source, we need a large impedance; by taking size into account, capacitors are optimum choices for implementation of large impedances. Therefore, instead of the resistors in the Howland’s solution we used capacitors with very small values. This implies the need to provide a DC pass for the current in the feedback loop of opamp UAmp2. Note that also in the DC regime, the output impedance must be high enough in order to have a good current source. Therefore we used the channel resistor of unbiased small NMOSs, QM1 and QM2 in order to provide very high value impedance and at the same time occupying small silicon area for delivering a DC pass for the feedback current. Such circuit is able to deliver a tuneable zero-mean white or pink-noise to the neuron.

\section{Spontaneous spiking activity of an individual silicon neuron stimulated by noise}
\begin{figure}[!ht]
\centering
\includegraphics[width=1\textwidth]{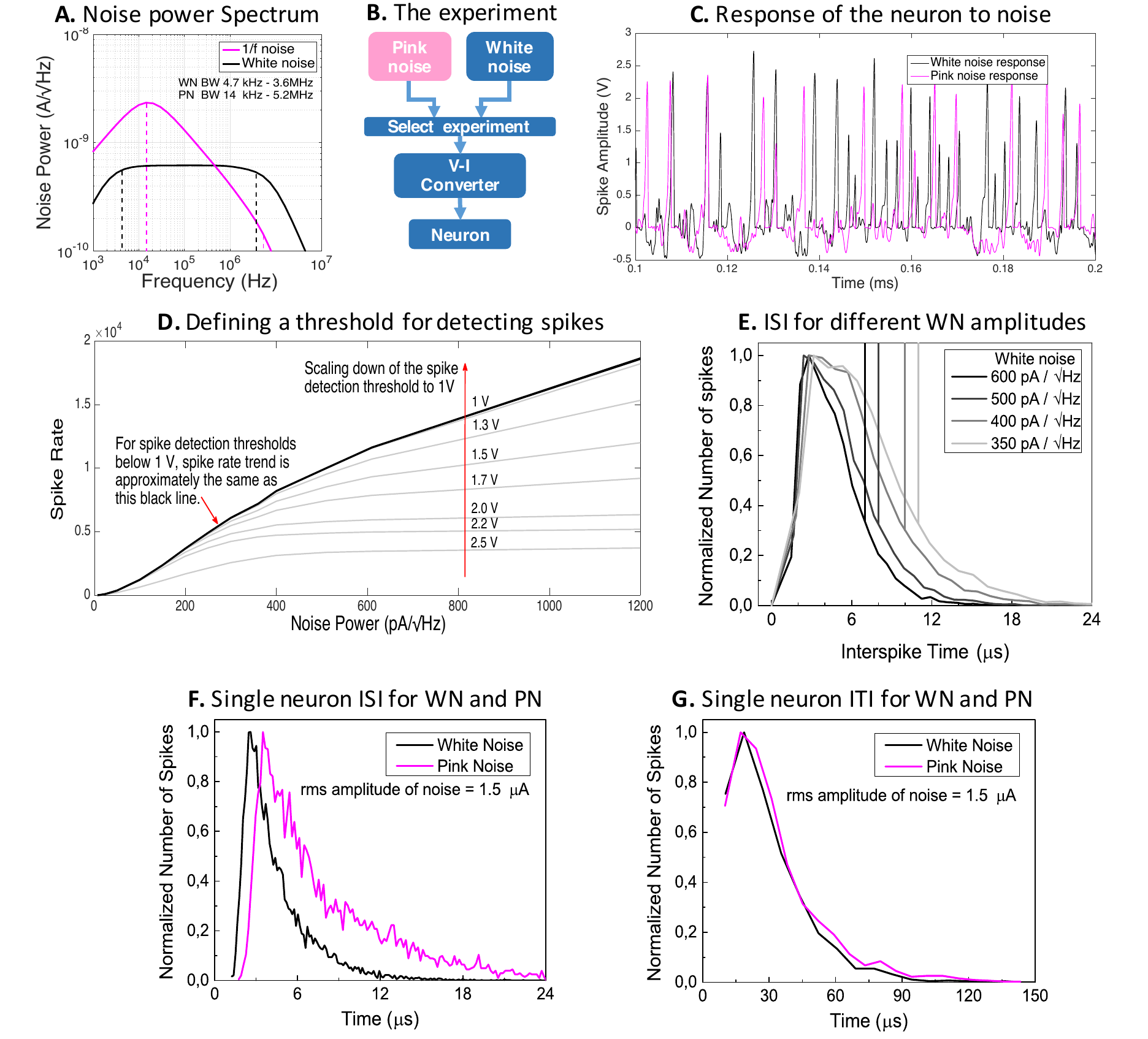}
\captionsetup{width=1.5\linewidth}
\caption{Injecting noise into an individual silicon neuron. A) The power spectrum of the white and pink-noise. Limitations on the noise bandwidth is determined by the circuit constraints. B) Flowchart of the simulation of a single neuron stimulated by noise. C) Response of the silicon neuron to white and noise input current with a rms amplitude of 1.5 $\mu A$. D) Determination of the threshold voltage to identify all the spikes induced by the noise injection. The counting stabilizes for threshold at 1 V. E) Inter-spike intervals (ISI) histogram  for the 50 ms transient simulation of the silicon neuron, applying a zero mean Gaussian white-noise to its input. Note that by increasing the amplitude of the noise the histogram is denser and shifted to the shorter time intervals. This is due to reduction of the relative refractory period of a neuron as a result of the increasing of the excitability of the neuron in case of stronger input stimulus. The vertical lines show the response of our silicon neuron in case of a constant input current with an amplitude equivalent to the noise rms values injected in 4 simulation in absence of noise. F) Comparison of the ISI of the spiking activity of the neuron stimulated by white and pink-noise, respectively. G) comparison of the inter-train intervals (ITI) histogram of the spiking activities of the neuron under white and pink-noise stimulations}
\label{f3}
\end{figure}

We now turn to the response of a single silicon neuron to injected input noise to address stochastic nature of the spiking activity, the inter-spike interval (ISI) distribution of the spiking activity and the relationship between input current noise amplitude and the spiking rate of the neuron, respectively.
Two transient simulations with the duration of 50 ms on the silicon neuron are performed using Cadence software and Spectre simulator where Gaussian white and pink-noise are injected to the neuron, respectively. The rms amplitude of the current noise in both cases is equally set to 1.5 $\mu A$. Such noise value causes the neuron to generate adequate spike rate in order to perform proper ISI and inter-train intervals (ITI) analysis. As a result, one can explore the features of the responses of the neuron to different noise spectrum signals and compare how the inter-spike intervals ISI and ITI are affected. Figure \ref{f3}A shows the power spectrum of the current noises employed to carry the simulation tuned so to cover the operating frequency of the silicon neuron (1MHz). Figure \ref{f3}C depicts the spiking activity of the silicon neuron in both simulations. 

ISI distribution of the spiking activity of a neuron helps quantifying the electrical properties of the neuron such as its refractory period, mean-firing rate and randomness of the spiking pattern, and therefore of the network \cite{stein2005neuronal}. Figure \ref{f3}D describes how the number of spike events varies upon increase of the amplitude of the input noise. Normally, the silicon neuron fires spikes with an amplitude of 2.5V when a deterministic current stimulates it. For detection of a spike, a threshold voltage above which the counter counts one spike has been determined. Figure \ref{f3}D includes the trend of the number of detected spikes versus injected input noise for different spike detection thresholds. Counting stabilizes for detection threshold voltage of 1 V. For a threshold less than 1V, random fluctuations would be considered as spike events which lead to faulty information.
Figure \ref{f3}E shows the ISI histogram for the 50 ms transient simulation of the silicon neuron, by applying a zero-mean Gaussian white-noise to its input. The simulation is repeated for 4 different amplitudes of the injected input white-noise ranging from 350 to 600 $pA/ \surd{Hz}$. By increasing the amplitude of the noise the histogram is denser and shifted to the shorter time intervals. The vertical line shows the natural period of spike activity of the neuron in case of a constant input current, with amplitude equivalent to the noise rms values injected in the simulations. A high number of spike events lay on the left side of their corresponding no--noise simulation results. Notice that the cross point of each histogram with its corresponding no--noise condition happens approximately at the same spike rate.
ISI and ITI of the spiking activities of the silicon neuron between the case of white and pink-noise stimuli are compared in Figure \ref{f3}E and \ref{f3}F. The flicker noise makes a longer tail on the histogram than the white-noise in agreement with other reports \cite{sobie2011neuron}. Furthermore, the histogram in case of white-noise is much denser than the one in the case of pink-noise. This implies that white-noise stimulus increases the excitability of a neuron within its natural refractory period with a higher rate in comparison to the pink-noise stimulus. In addition to analysis on the ISI distributions, we also plot the ITI histograms to see how different sources of noise affect the propagation of a train of spike events. For both white and pink-noise stimuli, the distribution of spiking activity is approximately concurrent. For the simulation of the correlation among the islands discussed in the next section, we adopt white noise who proves more effective to induce excitation of the neurons from the ISI point of view.

\section{Controlling the correlation of spiking activity of neuronal islands by tuning the interconnections}
With the aim of emulating and reverse engineering the test system topology constituted of four neuronal islands of biological neurons, we designed and simulated a network of spiking neurons comprising 64 silicon neurons and 1024 DPI synapses including a background white-noise. 
The case of an individual island comprising 16 silicon neurons and 256 synapses is first described. As anticipated, individual islands are stimulated with white-noise. Next, the increase of correlation of spiking activity of the four islands of neurons is controlled by tuning the number of the equivalent of interneurons connecting distinct neuronal islands, and the number of synapses excited by the same output interneuron. The observations are quantitatively accounted for by employing correlation coefficient matrix of stochastic events (Appendix for methods).

\subsection{Simulation of Islands of Silicon Neurons with Internal Low-Density Connectivity}
\begin{figure}[!ht]
\centering
\includegraphics[width=1\textwidth]{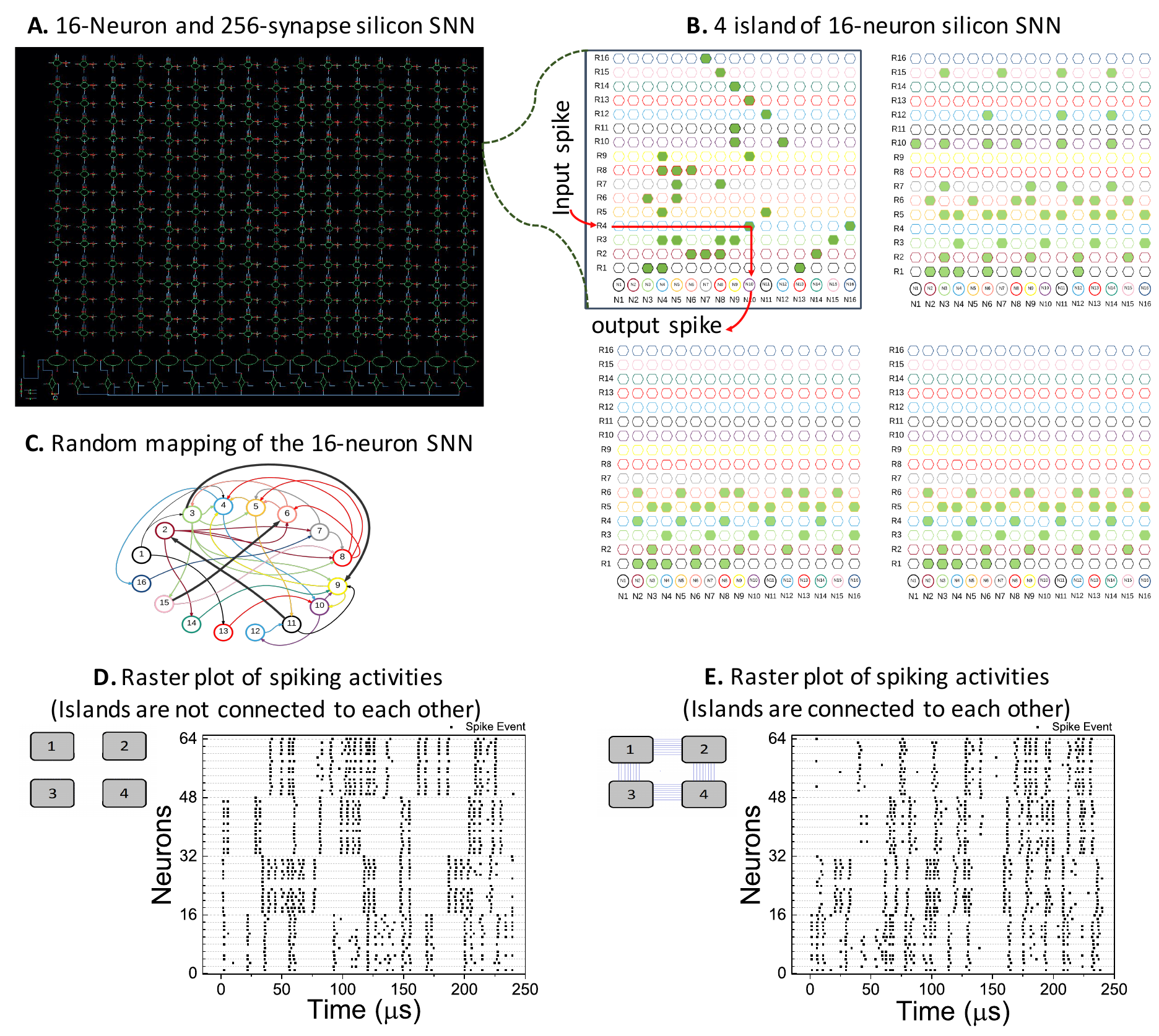}
\captionsetup{width=1.5\linewidth}
\caption{Injection of noise into a network of spiking neurons. A) Network of 16 neurons (ellipses on the bottom of the diagram) and 256 synapses (circles). For each neuron one V-I converter (diamond boxes, bottom of the diagram) has been utilized. B) Four islands of 16 neurons employed for simulation of measuring the activity-correlation by adding connections equivalent to interneurons. The neurons of the same island are connected to the same source of noise which is, in turn, uncorrelated to the noise sources of the other islands. 8\% of connections are inhibitory. C) A random topology realized on the 16-neuron network shown in part A. Thick black arrows represent inhibitory synapses while other arrows are excitatory synapses. D) Spiking activity of the neurons of the four island network excited by white-noise. The black dots show the spike event of a neuron. Neurons N1 to N16 are in first islands, from N17 to N32 in second island, from N33 to N48 in the third island and from N49 to N64 belong to the forth island. The spiking activity of neurons is correlated within the same island and uncorrelated between distinct islands. E) Spiking activity of the neurons of the case D when the islands are connected to each other through 8 distinct interneurons. The spiking activity of neurons of distinct islands becomes correlated.}
\label{f4}
\end{figure}

Figure \ref{f4}A represents a network of 16 silicon neurons and 256 reconfigurable synapses. In order to be consistent with networks of typical hybrid systems of real neurons on artificial solid state structures, we adopted a ratio between inhibitory and excitatory synapses in the range 5\% to 10\%. In our case such ratio is set to 8\%. The output of each neuron is connected to 16 synapses presented in a row in order to be able to connect to other 15 neurons and itself. The output of a neuron is connected directly to the input of each synapse. The output of the synapse is connected to the input nodes of the next neurons through a nMOS switch. By choosing whether the switch is closed or open, one can connect arbitrarily any neuron to the others. A Gaussian white-noise is applied through the V-I converter to each neuron. A single source of noise (noise of an on-chip resistor) distributes the noise to all neurons of an island in order to emulate local potential fluctuations affecting similarly all the neurons within the same neuronal island. Each island is solicited by a distinct noise generator, to emulate distinct regions with no common local potential fluctuations.

%Now that the 16-neuron network of spiking silicon neurons is implemented and is ready for being utilized, 
A random network topology (Figure \ref{f4}C) is configured on the 16-neuron network.
%to see how this network reacts to a Gaussian white-noise input applied to all of the neurons. As we observed in the experiment on the network of randomly connected biological neurons in Section 2, we expect to see highly correlated spiking activity of neurons regardless of the network topology since the same input pattern of noise is applied to every single neuron. 
We define four islands of such 16-neuron SNN as it is shown in Figure \ref{f4}B. Each island of neurons consist of different network topologies. In each island, 92\% of the synapses are excitatory and 8\% inhibitory which emulates a reasonable average ratio typical of realistic biological networks of spiking neurons.

\begin{figure}[!ht]
\centering
\includegraphics[width=1\textwidth]{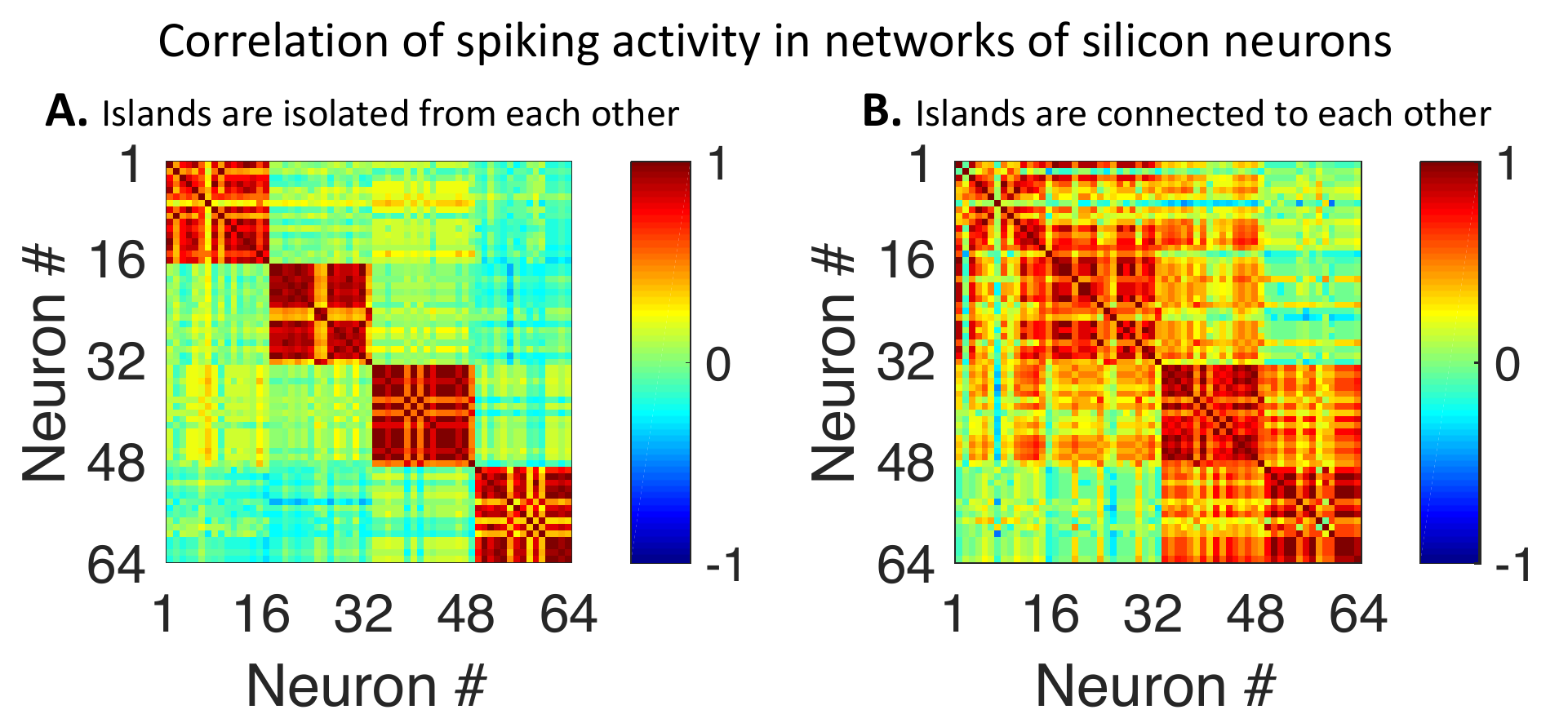}
\captionsetup{width=1.5\linewidth}
\caption{ Correlation coefficient matrices of spiking activity of networks of artificial neurons. The color-bar shows the amount of correlation coefficient from uncorrelated regions (blue) to highly correlated areas (dark red). A) Correlation coefficient matrix of the activity of network of silicon neurons where islands are not mutually connected. Spiking activity is highly synchronized inside each island but almost uncorrelated among different islands. B) Correlation coefficient matrix when island of silicon neurons are communicating with each other by 8 inter-connecting synapses in a ring topology.}
\label{f5}
\end{figure}

In Figure \ref{f4}B, circles and hexagons represent neurons and synapses respectively. The green synapses are activated. For instance, in the row number 4 (R4) of the Island 1 (top left), the 10th synapse is green, which indicates that neuron N4 stimulates neuron N10 by means of that synapse.

In the first island (top left) we implement a low-density connection in a random topology; on the contrary, in the second island (top right), we implement more connections between neurons, in order to demonstrate the influence of the variation of number of connections of each island on the correlation of the spiking activities.

We adopt the same random network topology on the third (bottom left) and forth (bottom right) islands, with the purpose of observing the variation of spiking activity of neurons when we inject uncorrelated sources of noise to the same network topology. As said, all neurons in each island are connected to a single source of noise. 

Our aim is to characterize the influence of spike trains generated by distinct regions and to observe how the correlation among spiking activity is created thanks to the propagation of the signaling on the surrounding network of the receiving neuron. The correlation propagates thanks to equivalent of interneurons and synchronization of spiking activity is rapidly achieved when few inter-connective synapses are employed.

In order to investigate the variation of correlation between the spiking activities in different islands we perform simulations by using the network represented in Figure \ref{f4}B.

In the first configuration, the islands of neurons are not mutually connected. By applying uncorrelated current-mode Gaussian white-noise with a power of 200 $pA/ \sqrt{Hz}$, to the islands for a simulation time of 240 μs, we obtain the spiking activity of the neurons represented in Figure \ref{f4}D. As a result, we obviously observe highly correlated firing activity only from neurons of the same island. 

In the second simulated configuration, islands of neurons are connected in a ring topology through interconnected synapses, such that the island 1 is connected to the island 2 by means of 8 interneurons via unidirectional synapses, in the same way, the island 2 is connected to island 4 , the 4 to the 3, and the 3 to the island 1. In such ring topology we achieved the spiking activity of the neurons represented in Figure \ref{f4}E, which looks correlated.

In order to assess quantitatively the correlation, we explore the correlation coefficient matrix. 
Figure \ref{f5} illustrates the correlation coefficient matrices of spiking activity of islands of silicon neurons in the mentioned simulations. In matrices, the correlation spectrum color-bar represents uncorrelated regions with blue and highly correlated regions with dark red.

When the islands are not connected (Figure \ref{f5}A), as a result of applying the same noise to individual islands and of their mutual connections, self-spiking activity of islands is highly correlated while, like in the case of biological neurons, neuron activity of different islands is not correlated.

When islands are connected to each other (Figure \ref{f5}B) by means of inter-connective synapses, activity of four islands coincides and therefore we observe universal highly correlated regions within the networks. We noticed that, in order to observe similar correlation to the biological case with only 1 connectons, 8 neuron-to-neuron connections are needed, suggesting that each neurite connects to more neurons within the same island. The condition of multiple synapses associated to the same neuron is developed in the next Section.

\subsection{Simulation of Islands of neurons with multiple synapses}
In order to achieve higher correlation between the island by adding only few neuron connections, like in the biological network of four islands, the simulated systems is modified in order to make the micro-connectome of the artificial interneurons more effective.

\begin{figure}[!h]
\centering
\includegraphics[width=1.0\textwidth]{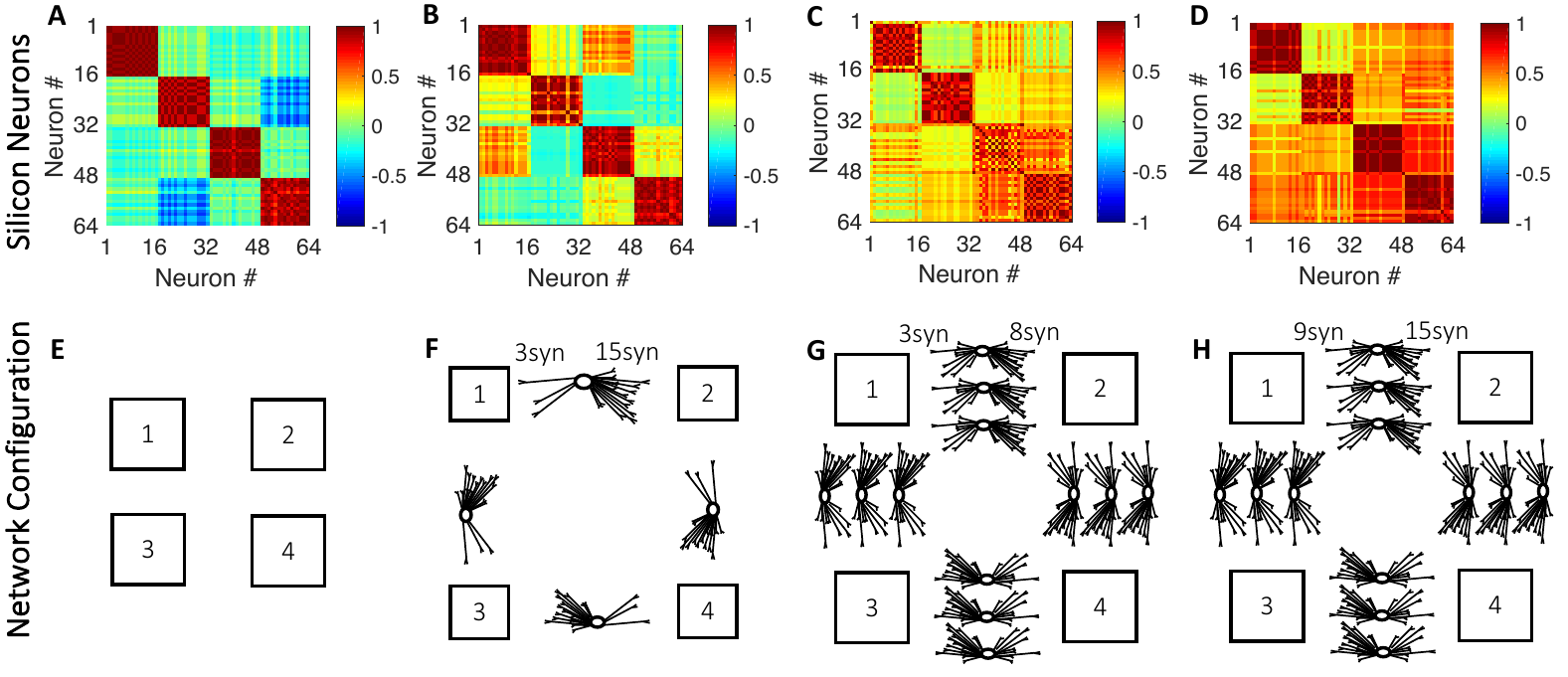}
\captionsetup{width=1.5\linewidth}
\caption{ Correlation coefficient matrices of spiking activity of networks of artificial neurons. In all graphs, the color-bar shows the amount of correlation coefficient from uncorrelated regions (blue) to highly correlated areas (dark red) Correlation activity in Islands of: A) silicon neurons isolated from each other.   B) silicon neurons with one interneuron connection in a ring topology, of kind ($3\times15$) C) silicon neurons with 3 interneuron connections ($3\times8$)  in a ring topology, D) silicon neurons with 3 connections  interconnectivity ($9\times15$) , E) represents silicon interneuron topology ($0\times0$), F) case of one connection per pair of islands linked as a ring, of kind ($3\times15$) G) case of three connections per pair of islands linked by ($3\times8$), which shows a correlation matrix similar to biological network with triple bond.
H) shows the limiting case of three connections of kind ($3\times15$).
}
\label{f6}
\end{figure}

Figure \ref{f6}A to \ref{f6}D, represents the correlation matrix of the activity of the neurons in cases of isolated islands, which refer to the connections represented in \ref{f6}E to \ref{f6}H. The case A-E ($0\times0$) provides the background of the correlation when there are no connection. The case B-F ($3\times15$) represents a limiting case when one neuron solicits all the neurons in the next islands. It is worth noting that even if all the neurons of the next island are solicited, the correlation does not saturate among distinct islands. The case C-G with three connections of kind ($3\times8$) shows much stronger correlation even if the synapses to input neurons on the next islands are cut of a half, thanks to the triple excitation connection. Such condition provides a result which is closer to the 3N connection of biological islands, namely a good correlation with only three connections, if compared to the individual point-to-point connections discussed previously. Finally the case D-H of three ($9\times15$) connected neurons represents again a limiting case where all the neurons of the next islands are fed, which exhibits even higher (but not saturated despite the connection to all the neurons of the next island) correlation. The lack of saturation may be explained by the strong hypothesis set by injecting uncorrelated noise to the four islands, which in turn may become a limiting boundary condition to the system, and does not account for independent potential fluctuations of the interneurons. Such findings suggest that in the biological islands the spontaneous connections of the interneurons with the neurons of the islands, in spite of the process and the methods, are very likely and very efficient. As a finel remark about the level of smaller correlation among the islands of the silicon network compared to biological network, there is some effect also ascribed to smaller populations of the silicon islands, and to higher degree of connections between interneurons and the islands. 

     ~

\section{Conclusions}

To conclude, we compared a network of four randomly connected biological and silicon 
islands, equipped of additional interneurons to enable communication between distinct islands, and therefore push spontaneous firing into higher correlation. We proposed an original and robust implementation of artificial silicon neurons by designing compact and low power consuming electronic CMOS microcircuits, including both pink and white noise generators.  We employed noise to activate the spontaneous firing activity of the neurons and we characterized the statistics of the response of individual neurons. Finally, the simulations of a ring of four noise--activated silicon islands and its comparison with its twin made of real neurons enabled us to reverse engineer the nature of the connections formed between the real neurons of the biological system. The results suggest that several synapses of each interneuron employed to connect distinct islands are required to grant the high correlation experimentally observed.

\section*{Acknowledgments}
E.P. acknowledges JSPS Fellowship, the Hokkaido University and the Short Term Mobility Program 2016 of CNR.

\section*{Appendix}

\textbf{Correlation coefficient matrix calculation}

We aim to mathematically characterize the influence of spike trains generated by the neurons of each island and observe how the correlation among spiking activity is created. For this purpose we use Pearson correlation coefficient of two random variables which is a linear measure of their dependence \cite{stuart1968advanced}. If each of the variables has N samples, the Pearson correlation coefficient is defined as:
\begin{equation}
	\label{eq3}
	\rho (A,B) = \frac{1}{N-1} \sum_{i=1}^{N}(\frac{A_i - \mu _A}{\sigma _A})(\frac{B_i - \mu _B}{\sigma _B}),
\end{equation}

where $\mu A$ and $\sigma A$ are the mean and standard deviation of $A$, respectively, and $\mu B$ and $\sigma B$ are the mean and standard deviation of $B$. Correlation coefficient can be written in terms of covariance of $A$ and $B$:
\begin{equation}
	\label{eq4}
	\rho (A,B) = \frac{A,B}{\sigma _A \sigma _B}
\end{equation}
The correlation coefficient matrix of two random variables for all pairs of variables as the following:
\begin{equation}
R = 
 \begin{pmatrix}
  \rho (A,A) & \rho (A,B) \\
  \rho (B,A) & \rho (B,B) \\
 \end{pmatrix}
\end{equation}
$A$ and $B$ are directly correlated to themselves thus the diagonal entries of the matrix are 1. The matrix is symmetric such that $\rho (A,A) = \rho (A,B)$
\begin{equation}
	R = 
 \begin{pmatrix}
  1 & \rho (A,B) \\
  \rho (B,A) & 1 \\
 \end{pmatrix}
\end{equation}

Since spiking activities of the neurons inside the 64 neuron network are random variables we can simply build their correlation coefficient matrix $(64 \times 64)$ in order to find their dependence, quantitatively. For calculation of the correlation coefficient matrix of the spiking activity of neurons we utilize MATLAB function corrcoef(X). It returns the matrix of correlation coefficients for X, where the columns of X represent random variables and the rows represent the samples.
%\section*{References}

\bibliography{references}

\end{document}